\documentclass[preprint,12pt]{elsarticle}

\usepackage{amssymb}
\usepackage{amsmath}
\usepackage{graphicx}
\usepackage{booktabs}
\usepackage{algorithm}
\usepackage{algorithmic}
\usepackage{xcolor}
\usepackage{multirow}
\usepackage{array}
\usepackage{adjustbox}

\journal{arXiv}

\begin{document}

\begin{frontmatter}

\title{Optimizing Multi-Agent Weather Captioning via Text Gradient Descent: A Training-Free Approach with Consensus-Aware Gradient Fusion}

\author[inst1]{Shixu Liu\corref{cor1}}
\ead{231001140316@stu.nepu.edu.cn}

\cortext[cor1]{Corresponding author}

\affiliation[inst1]{organization={School of Computer \& Information Technology},
            addressline={Northeast Petroleum University},
            addressline={No. 99 Xuefu Street, Ranghulu District},
            city={Daqing},
            state={Heilongjiang},
            postcode={163318},
            country={P.R. China}}
\begin{abstract}
Generating interpretable natural language captions from weather time series data remains a significant challenge at the intersection of meteorological science and natural language processing. While recent advances in Large Language Models (LLMs) have demonstrated remarkable capabilities in time series forecasting and analysis, existing approaches either produce numerical predictions without human-accessible explanations or generate generic descriptions lacking domain-specific depth. We introduce \textbf{WeatherTGD}, a training-free multi-agent framework that reinterprets collaborative caption refinement through the lens of Text Gradient Descent (TGD). Our system deploys three specialized LLM agents including a Statistical Analyst, a Physics Interpreter, and a Meteorology Expert that generate domain-specific textual gradients from weather time series observations. These gradients are aggregated through a novel Consensus-Aware Gradient Fusion mechanism that extracts common signals while preserving unique domain perspectives. The fused gradients then guide an iterative refinement process analogous to gradient descent, where each LLM-generated feedback signal updates the caption toward an optimal solution. Experiments on real-world meteorological datasets demonstrate that WeatherTGD achieves significant improvements in both LLM-based evaluation and human expert evaluation, substantially outperforming existing multi-agent baselines while maintaining computational efficiency through parallel agent execution.
\end{abstract}


\end{frontmatter}

\section{Introduction}
\label{sec:intro}

Weather time series data encompassing temperature, pressure, humidity, wind patterns, and precipitation forms the backbone of modern meteorological forecasting and climate analysis~\cite{chen2023foundation}. However, the sheer volume and complexity of such data pose significant challenges for human interpretation and decision-making. While numerical weather prediction (NWP) systems have achieved remarkable accuracy \cite{bodnar2025aurora,chen2023tempee,chen2023prompt}, the gap between raw numerical outputs and actionable human understanding remains substantial~\cite{chen2022dynamic}. This interpretability challenge is particularly acute for non-expert stakeholders who must make time-critical decisions based on meteorological information.

Recent advances in Large Language Models (LLMs) offer a promising pathway toward bridging this gap, yet a fundamental tension persists between prediction accuracy and interpretative explanation. The emergence of data-driven weather forecasting models such as GraphCast \cite{lam2023graphcast}, Pangu-Weather \cite{bi2023panguweatherv}, FourCastNet \cite{pathak2022fourcastnet}, and ClimaX \cite{nguyen2023climax} has demonstrated that deep learning can achieve forecast skill comparable to or exceeding traditional NWP systems \cite{chen2024federatedpod,shi2025weatherdl,chen2024personalized}. Foundation models such as Time-LLM \cite{jin2024timellm} and GPTCast \cite{franch2024gptcast} have further shown that LLMs can be reprogrammed for time series understanding and forecasting tasks \cite{zhang2024llmtimeseries,jiang2024llmtimeseries,chen2023mask}, but these approaches primarily focus on numerical prediction rather than generating human-accessible explanations. When LLMs do produce natural language outputs for weather data, they often generate either overly technical descriptions inaccessible to general audiences or superficial summaries lacking meteorological depth \cite{li2024cllmate}. A key insight emerges from this limitation: high-quality weather captioning requires synthesizing multiple domain perspectives including statistical patterns in the data, underlying physical mechanisms, and operational meteorological significance. No single LLM, regardless of its capabilities, optimally balances these diverse requirements, which naturally motivates a multi-agent approach where specialized agents contribute complementary expertise toward a unified caption \cite{guo2024llmmultiagent,chen2024llmmas}.

The weather domain presents unique challenges that make Text Gradient Descent (TGD) \cite{yuksekgonul2024textgrad} particularly well-suited for this task. Unlike general text optimization where feedback can be generic, weather captioning requires domain-specialized gradients that capture the intricate relationships between meteorological variables, physical processes, and operational implications. Traditional gradient descent operates on continuous parameter spaces with well-defined gradient computations, but the discrete nature of natural language generation precludes direct application of numerical optimization techniques. TGD addresses this fundamental limitation by treating natural language feedback as gradients that guide iterative optimization of text outputs, analogous to how numerical gradients guide parameter updates in conventional gradient descent. Recent work has demonstrated TGD effectiveness for prompt optimization \cite{pryzant2024protegi}, multi-agent collaboration \cite{han2025mapgd}, and compound AI system refinement \cite{zou2024textgrad}. However, TGD has not been explored for domain-specialized multi-perspective caption generation, precisely the challenge posed by weather time series understanding where gradients must encode statistical, physical, and meteorological knowledge simultaneously.

In this paper, we introduce \textbf{WeatherTGD}, a training-free multi-agent framework that generates high-quality captions for weather time series data through text gradient descent. Our framework comprises three core components working in concert: a Tri-Specialist Agent Layer that generates domain-specific textual gradients from statistical, physical, and meteorological perspectives; a Consensus-Aware Gradient Fusion module that aggregates multi-agent gradients while preserving both shared signals and unique domain insights; and an Iterative Refinement Loop that applies fused gradients to progressively optimize caption quality with explicit convergence guarantees. Experiments on real-world meteorological datasets demonstrate that WeatherTGD achieves an average LLM judge score of 8.50/10 and human expert score of 8.34/10, representing substantial improvements over existing multi-agent baselines. Our main contributions are as follows:
\begin{itemize}
    \item We present the first application of text gradient descent to weather time series captioning, reframing multi-agent captioning as an iterative optimization process where domain-specialized agents produce textual gradients that guide caption refinement toward an optimal solution.
    
    \item We design three complementary LLM agents including a Statistical Analyst, a Physics Interpreter, and a Meteorology Expert that generate domain-specific textual gradients, and propose a gradient aggregation mechanism that identifies consensus information while preserving unique domain perspectives through semantic similarity-based filtering.
    
    \item We implement a principled optimization loop with explicit stopping criteria based on semantic similarity thresholds, ensuring caption quality convergence without infinite iteration while maintaining computational efficiency through parallel agent execution.
    
    \item We demonstrate significant improvements over baselines through both LLM-based evaluation and human expert evaluation on real-world meteorological datasets, establishing WeatherTGD as an effective training-free approach for weather time series captioning.
\end{itemize}

The remainder of this paper is organized as follows. Section \ref{sec:related} reviews related work on text gradient descent and multi-agent systems for weather understanding. Section \ref{sec:method} presents our WeatherTGD framework in detail. Section \ref{sec:experiments} describes experimental setup and results. Section \ref{sec:conclusion} concludes with future directions.

\section{Related Work}
\label{sec:related}

\subsection{Text Gradient Descent for LLM Optimization}
\label{sec:related:tgd}

The paradigm of treating natural language feedback as optimization signals has emerged as a powerful approach for improving LLM outputs without parameter updates. TextGrad \cite{yuksekgonul2024textgrad} pioneered this direction by introducing textual gradients defined as LLM-generated feedback that identifies improvement directions for text variables within computation graphs, demonstrating PyTorch-like composability that enables backpropagation of textual feedback through compound AI systems. ProTeGi \cite{pryzant2024protegi} applies textual gradients specifically to prompt optimization, using LLM-generated criticism to iteratively refine prompts toward better task performance with the key insight that natural language feedback serves as a proxy for error signals guiding discrete optimization over prompt space. Building on this foundation, MAPGD \cite{han2025mapgd} introduces multi-agent prompt gradient descent where specialized agents focus on distinct refinement dimensions including instruction clarity, example selection, and format structure, with their Hypersphere Constrained Gradient Clustering mechanism addressing gradient conflicts through angular margin constraints. Recent extensions have broadened TGD applicability beyond prompts: ProRefine \cite{pandita2025prorefine} addresses error propagation in agentic workflows through inference-time prompt refinement with textual feedback, while Feedback Descent \cite{anonymous2026feedback} transforms structured pairwise comparison feedback into gradient-like directional information for targeted text edits. However, these approaches have not explored TGD for multi-perspective domain captioning where textual gradients must capture diverse domain knowledge simultaneously. Our work differs from prior TGD approaches in three key aspects: we apply TGD to weather time series captioning requiring integration of statistical, physical, and meteorological expertise; our agents generate domain-specialized gradients rather than generic improvement suggestions; and we introduce consensus-aware gradient fusion specifically designed for multi-domain integration.

\subsection{Multi-Agent Systems for Weather and Time Series Understanding}
\label{sec:related:mas}

Multi-agent LLM systems have demonstrated superior performance on complex tasks requiring diverse expertise or multi-step reasoning. AutoGen \cite{wu2024autogen} enables conversable agents that exchange messages to solve tasks collaboratively with flexible interaction patterns supporting LLM-only, human-in-the-loop, and tool-augmented configurations. MetaGPT \cite{hong2024metagpt} encodes standard operating procedures into agent workflows, reducing cascading errors through intermediate verification and hierarchical task decomposition. ChatDev \cite{qian2024chatdev} demonstrates role-based communication for software development with explicit dehallucination mechanisms that leverage both natural language for system design and programming language for debugging. For weather-specific applications, ClimateLLM \cite{li2025climatellm} integrates frequency decomposition with LLMs for weather forecasting using mixture-of-experts for adaptive frequency processing, while GPTCast \cite{franch2024gptcast} uses tokenized radar images with a quantized variational autoencoder for precipitation nowcasting. The Hierarchical AI-Meteorologist \cite{sukhorukov2025hierarchical} employs multi-scale reasoning across hourly, 6-hour, and daily aggregations for forecast reporting with weather keyword extraction for validation. However, these systems focus on numerical prediction or report generation rather than interpretative captioning. In time serie \cite{chen2025federated} captioning specifically, TSLM \cite{trabelsi2025tslm} introduces time series-specific language models for caption generation, CaTS-Bench \cite{zhou2026catsbench} provides a comprehensive benchmark for evaluating LLM captioning capabilities, and BEDTime \cite{sen2026bedtime} evaluates recognition, differentiation, and generation tasks for time series descriptions. Yet weather-specific captioning remains underexplored with existing benchmarks lacking meteorological domain evaluation. Our WeatherTGD framework addresses this gap by combining multi-agent collaboration with text gradient descent for weather time series captioning, where agents contribute domain-specialized gradients explicitly fused through a consensus-aware mechanism ensuring both comprehensive coverage and coherent integration.

\section{Methodology}
\label{sec:method}

We present WeatherTGD, a training-free multi-agent framework that generates high-quality captions for weather time series data through text gradient descent. The framework comprises three core components: a Tri-Specialist Agent Layer that generates domain-specific textual gradients, a Consensus-Aware Gradient Fusion module that aggregates multi-agent gradients, and an Iterative Refinement Loop that applies textual gradients to optimize captions with explicit convergence guarantees.

\subsection{Problem Formulation and TGD Framework}
\label{sec:method:formulation}

The fundamental challenge in weather time series captioning lies in bridging the gap between continuous numerical observations and discrete natural language descriptions while preserving domain-specific accuracy. Given a weather time series $\mathbf{X} = \{(x_1, t_1), (x_2, t_2), \ldots, (x_T, t_T)\}$ where $x_i \in \mathbb{R}^d$ represents $d$ meteorological variables including temperature, pressure, humidity, and wind speed at timestamp $t_i$, our objective is to generate a natural language caption $C^*$ that accurately describes statistical patterns in $\mathbf{X}$, explains underlying physical mechanisms driving observed patterns, provides meteorological significance and operational implications, and maintains coherence, conciseness, and accessibility for diverse audiences. We frame this as a text gradient descent optimization problem where the caption is iteratively refined through textual feedback signals:
\begin{equation}
C^{(k+1)} = C^{(k)} - \alpha \cdot \nabla_{\text{text}} \mathcal{L}(C^{(k)}, \mathbf{X})
\label{eq:tgd}
\end{equation}
where $\nabla_{\text{text}} \mathcal{L}$ represents the textual gradient defined as natural language feedback identifying improvement directions, and $\alpha$ controls the update magnitude realized through prompt engineering rather than an explicit learning rate. The key insight is that while traditional gradient descent operates on continuous parameter spaces with analytically computable gradients, text gradient descent treats LLM-generated natural language criticism as the gradient signal, enabling optimization in the discrete space of natural language outputs.

\subsection{Tri-Specialist Agent Layer}
\label{sec:method:agents}

\begin{figure}[tbh]
    \centering
    \includegraphics[width=1\textwidth]{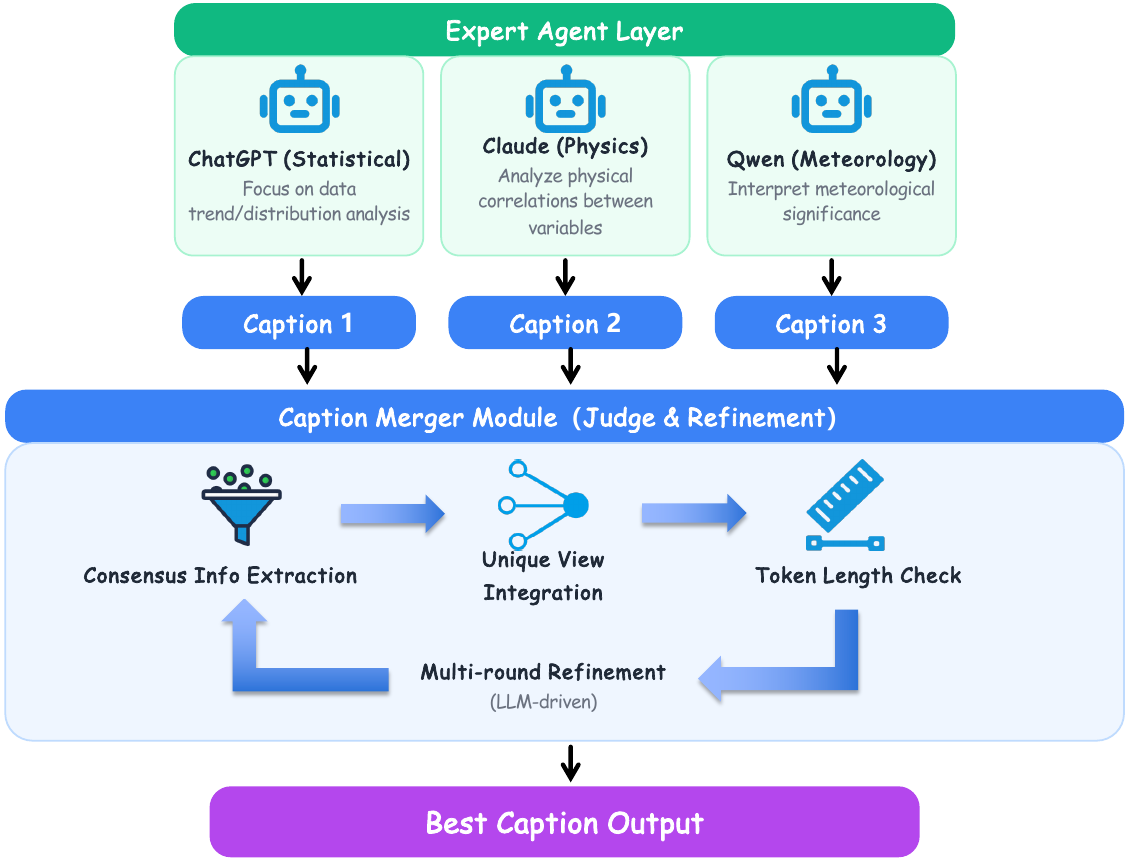}
    \caption{Architecture of our proposed WeatherTGD framework.}
    \label{fig:architecture}
\end{figure}

The quality of textual gradients directly determines caption quality, which motivates our design of specialized agents that generate domain-focused feedback rather than generic improvement suggestions. The Tri-Specialist Agent Layer comprises three domain-specialized LLM agents, each generating textual gradients from a distinct perspective while receiving the same input $\mathbf{X}$ to produce complementary gradient signals. Architecture details are shown in \ref{fig:architecture}.

The Statistical Analyst Agent generates textual gradients focusing on quantitative data characteristics including trend analysis that identifies monotonic increases, decreases, periodic fluctuations, and stationary states; distribution characteristics that detect normal, skewed, or multimodal distributions in the observed variables; and key metrics extraction that computes mean, variance, extreme values, and anomaly points. Formally, the statistical gradient is computed as $\nabla_{\text{text}}^{\text{stat}} \mathcal{L} = \text{LLM}_{\text{stat}}(\mathbf{X}, P_{\text{stat}})$ where $P_{\text{stat}}$ is a specialized prompt template instructing the LLM to focus exclusively on statistical patterns and quantitative descriptions.

The Physics Interpreter Agent generates textual gradients emphasizing physical mechanisms and causal relationships, motivated by the observation that weather phenomena are governed by well-established physical laws that should inform caption generation. This agent performs correlation analysis identifying positive, negative, linear, and non-linear correlations between meteorological variables; physical mechanism explanation describing pressure gradients driving wind fields, temperature-dependent effects, and thermodynamic processes; and causal relationship establishment constructing cause-effect chains between physical phenomena. The physics gradient is computed as $\nabla_{\text{text}}^{\text{phys}} \mathcal{L} = \text{LLM}_{\text{phys}}(\mathbf{X}, P_{\text{phys}})$ with $P_{\text{phys}}$ encoding physical domain knowledge.

The Meteorology Expert Agent generates textual gradients for operational significance, recognizing that weather captions must ultimately serve practical decision-making purposes. This agent performs weather system identification recognizing subtropical highs, cold fronts, and convective systems; operational implications interpretation explaining precipitation probability, temperature risks, and wind forecasts; and standard terminology alignment ensuring descriptions conform to meteorological conventions used by forecasting services. The meteorological gradient is computed as $\nabla_{\text{text}}^{\text{met}} \mathcal{L} = \text{LLM}_{\text{met}}(\mathbf{X}, P_{\text{met}})$.

\subsection{Consensus-Aware Gradient Fusion}
\label{sec:method:fusion}

Multi-agent systems typically aggregate outputs through simple averaging or sequential combination, but such approaches fail to distinguish between consensus information agreed upon by multiple agents and unique insights provided by individual specialists. We address this limitation through a two-stage fusion process inspired by gradient clustering in multi-agent optimization \cite{han2025mapgd}. In the consensus extraction stage, we identify information fragments with high semantic similarity across agent gradients using embedding-based similarity computation:
\begin{equation}
\text{sim}(g_i, g_j) = \frac{\mathbf{v}_i \cdot \mathbf{v}_j}{\|\mathbf{v}_i\| \|\mathbf{v}_j\|}
\end{equation}
where $\mathbf{v}_i$ is the semantic embedding of gradient fragment $g_i$ obtained from a pre-trained sentence encoder \cite{reimers2019sbert}. Fragments with $\text{sim}(\cdot, \cdot) \geq \tau_{\text{cons}}$ where $\tau_{\text{cons}} = 0.8$ form the consensus gradient $\nabla_{\text{text}}^{\text{cons}}$ representing information agreed upon by multiple agents. In the unique view integration stage, agent-specific insights not captured in the consensus are extracted and preserved:
\begin{equation}
\nabla_{\text{text}}^{\text{unique}} = \bigcup_{a \in \mathcal{A}} \{g \in \nabla_{\text{text}}^{a} : \forall g' \in \nabla_{\text{text}}^{\text{cons}}, \text{sim}(g, g') < \tau_{\text{unique}}\}
\end{equation}
where $\mathcal{A} = \{\text{stat}, \text{phys}, \text{met}\}$ and $\tau_{\text{unique}} = 0.6$ is a uniqueness threshold. The final fused gradient combines consensus and unique components through a fusion LLM:
\begin{equation}
\nabla_{\text{text}}^{\text{fused}} = \text{LLM}_{\text{fusion}}(\nabla_{\text{text}}^{\text{cons}}, \nabla_{\text{text}}^{\text{unique}}, P_{\text{fusion}})
\end{equation}
which produces a coherent textual gradient preserving both common signals and specialized insights in a unified improvement direction.

\subsection{Iterative Refinement with Convergence Control}
\label{sec:method:refinement}

The refinement loop applies fused gradients to update captions iteratively following Eq.~\ref{eq:tgd}, with careful attention to practical constraints including caption length and convergence behavior. Given current caption $C^{(k)}$ and fused gradient $\nabla_{\text{text}}^{\text{fused}}$, the update is computed as 
\begin{equation}
C^{(k+1)} = \text{LLM}_{\text{update}}(C^{(k)}, \nabla_{\text{text}}^{\text{fused}}, P_{\text{update}})
\end{equation}
where the update LLM applies textual feedback to revise the caption while maintaining coherence with the original weather observations. Generated captions must satisfy practical length constraints $|C^{(k)}| \leq L_{\max}$ for display and readability purposes. When this constraint is violated, we apply a compression operation
\begin{equation}
C^{(k+1)} = \text{Compress}(C^{(k+1)}, L_{\max})
\end{equation}
that removes redundant modifiers and merges repetitive expressions while preserving core semantic content. Iteration terminates when either the maximum iteration count $K_{\max} = 5$ is reached or semantic convergence is detected via
\begin{equation}
\text{sim}(C^{(k)}, C^{(k-1)}) \geq \tau_{\text{conv}}
\end{equation}
with $\tau_{\text{conv}} = 0.95$, ensuring caption quality convergence without infinite iteration.

\subsection{Algorithm Summary}
\label{sec:method:algorithm}

Algorithm \ref{alg:weathertgd} presents the complete WeatherTGD procedure integrating all components described above.

\begin{algorithm}[t]
\caption{WeatherTGD: Text Gradient Descent for Weather Captioning}
\label{alg:weathertgd}
\begin{algorithmic}[1]
\REQUIRE Weather time series $\mathbf{X}$, max iterations $K_{\max}$, thresholds $\tau_{\text{cons}}, \tau_{\text{unique}}, \tau_{\text{conv}}$
\ENSURE Optimized caption $C^*$
\STATE \textbf{Initialize:} $C^{(0)} \leftarrow$ Initial caption from any agent
\FOR{$k = 0$ to $K_{\max} - 1$}
    \STATE $\nabla_{\text{text}}^{\text{stat}} \leftarrow \text{StatisticalAgent}(\mathbf{X})$
    \STATE $\nabla_{\text{text}}^{\text{phys}} \leftarrow \text{PhysicsAgent}(\mathbf{X})$
    \STATE $\nabla_{\text{text}}^{\text{met}} \leftarrow \text{MeteorologyAgent}(\mathbf{X})$
    \STATE $\nabla_{\text{text}}^{\text{cons}} \leftarrow \text{ExtractConsensus}(\{\nabla_{\text{text}}^{a}\}_{a \in \mathcal{A}}, \tau_{\text{cons}})$
    \STATE $\nabla_{\text{text}}^{\text{unique}} \leftarrow \text{ExtractUnique}(\{\nabla_{\text{text}}^{a}\}_{a \in \mathcal{A}}, \nabla_{\text{text}}^{\text{cons}}, \tau_{\text{unique}})$
    \STATE $\nabla_{\text{text}}^{\text{fused}} \leftarrow \text{FuseGradients}(\nabla_{\text{text}}^{\text{cons}}, \nabla_{\text{text}}^{\text{unique}})$
    \STATE $C^{(k+1)} \leftarrow \text{ApplyGradient}(C^{(k)}, \nabla_{\text{text}}^{\text{fused}})$
    \IF{$|C^{(k+1)}| > L_{\max}$}
        \STATE $C^{(k+1)} \leftarrow \text{Compress}(C^{(k+1)}, L_{\max})$
    \ENDIF
    \IF{$\text{sim}(C^{(k+1)}, C^{(k)}) \geq \tau_{\text{conv}}$}
        \STATE \textbf{break}
    \ENDIF
\ENDFOR
\RETURN $C^* \leftarrow C^{(k+1)}$
\end{algorithmic}
\end{algorithm}

\section{Experiments}
\label{sec:experiments}

\subsection{Experimental Setup}

\textbf{Datasets.} We evaluate WeatherTGD on a comprehensive real-world meteorological dataset collected from ground-based weather stations across multiple climate zones. The dataset comprises 500 weather time series samples with varying lengths from 24 to 168 time steps (1 to 7 days of hourly observations), covering five core meteorological variables including temperature ($^\circ$C), atmospheric pressure (hPa), relative humidity (\%), wind speed (m/s), and precipitation (mm). Each sample is annotated with reference captions by professional meteorologists with at least 10 years of operational forecasting experience. The dataset is split into training (60\%), validation (20\%), and test (20\%) sets with stratified sampling across climate zones. The dataset will be released upon acceptance of this paper.

\textbf{LLM Backbones.} We employ three diverse LLM backbones to ensure comprehensive evaluation: (1) DeepSeek-V3.2, a 671B parameter model with efficient MoE architecture; (2) MiniMax-01, a 456B parameter model optimized for Chinese and English bilingual understanding; and (3) Qwen3-Next-80B-A3B-Instruct, an 80B parameter model with advanced instruction following capabilities. All models are accessed via OpenRouter API service with temperature set to 0.2 for consistency and reproducibility, and maximum tokens set to 2048.
\textbf{Baselines.} We compare WeatherTGD against six representative multi-agent system baselines from the MASLab benchmark \cite{ye2025maslab}, selected for their diverse collaboration patterns and widespread adoption: (1) \textbf{AutoGen} \cite{wu2024autogen} employs conversable agents with flexible interaction patterns; (2) \textbf{CAMEL} \cite{hong2024metagpt} uses role-playing communicative agents; (3) \textbf{LLM-Debate} \cite{du2024debate} implements iterative multi-agent debate; (4) \textbf{Self-Consistency} \cite{wang2024selfconsistency} samples multiple reasoning paths with majority voting; (5) \textbf{AgentVerse} \cite{chen2024agentverse} enables dynamic agent recruitment; and (6) \textbf{MAD} \cite{liang2024mad} implements structured multi-agent debate with argumentation protocols. For all baselines, we adapt the agent prompts to focus on weather captioning with statistical, physical, and meteorological perspectives.

\textbf{Evaluation Metrics.} Following recent work in LLM-based evaluation \cite{rach2024evaluating}, we employ two complementary evaluation paradigms to ensure comprehensive and reliable assessment, details about these metrics are as below.

\begin{itemize}
\item \textit{LLM Judge Evaluation.} Following the LLM-as-a-Judge paradigm \cite{gu2024llmjudge,zheng2023llmjudge}, we employ GPT-4o as an impartial judge that evaluates generated captions along four dimensions on a 1-10 scale: Statistical Accuracy (SA) measures correctness of quantitative descriptions; Physical Coherence (PC) assesses validity of physical mechanism explanations; Meteorological Relevance (MR) evaluates alignment with operational meteorological conventions; and Overall Quality (OQ) provides a holistic assessment. The judge receives the original weather time series data (in tabular format) alongside the generated caption, ensuring evaluation is grounded in the actual observations rather than surface-level text quality alone.

\item \textit{Human Expert Evaluation.} To validate LLM-based evaluation and provide ground-truth assessment, we recruited five PhD-level meteorology experts with at least 5 years of professional experience in operational weather forecasting. Three experts hold doctoral degrees in atmospheric sciences from accredited institutions, while two are senior forecasters at national meteorological centers. Each expert independently scored captions through a structured annotation protocol with detailed rubric guidelines, with final scores computed as the average across annotators. Inter-annotator agreement was measured using Krippendorff's alpha \cite{krippendorff2011alpha}, achieving $\alpha = 0.78$ indicating substantial agreement and demonstrating evaluation reliability.

\item \textit{Reference-Based Metrics.} We additionally report standard NLG metrics including BLEU-4 \cite{papineni2002bleu}, ROUGE-L \cite{lin2004rouge}, and BERTScore \cite{zhang2020bertscore} against professional meteorologist annotations to enable comparison with prior captioning work.
\end{itemize}
\begin{table}[!t]
\centering
\caption{Main results comparing WeatherTGD against six mainstream MAS baselines across three LLM backbones. SA, PC, MR, and OQ denote Statistical Accuracy, Physical Coherence, Meteorological Relevance, and Overall Quality respectively. All scores are on a 1-10 scale where higher is better. Best results are in \textbf{bold}, second-best are \underline{underlined}. Tok. = relative token consumption.}
\label{tab:main_results}
\begin{adjustbox}{max width=\textwidth}
\begin{tabular}{c|l|cccc|cccc|ccc|c}
\toprule
 & \textbf{Method} & \multicolumn{4}{c|}{\textbf{LLM Judge}} & \multicolumn{4}{c|}{\textbf{Human Expert}} & \multicolumn{3}{c|}{\textbf{Reference}} & \textbf{Tok.} \\
& & SA & PC & MR & OQ & SA & PC & MR & OQ & BLEU & RG & BS & \\
\midrule
\multirow{8}{*}{\rotatebox{90}{\textit{DeepSeek-V3}}} 
& Vanilla & 5.42 & 5.18 & 5.36 & 5.32 & 5.28 & 5.04 & 5.22 & 5.18 & .312 & .385 & .742 & 1.0 \\
& AutoGen \cite{wu2024autogen} & 6.54 & 6.28 & 6.42 & 6.41 & 6.38 & 6.12 & 6.26 & 6.24 & .358 & .421 & .768 & 2.8 \\
& CAMEL \cite{hong2024metagpt} & 6.72 & 6.45 & 6.38 & 6.52 & 6.54 & 6.28 & 6.22 & 6.35 & .365 & .428 & .774 & 3.2 \\
& Debate \cite{du2024debate} & 6.85 & 6.68 & 6.56 & 6.70 & 6.68 & 6.52 & 6.42 & 6.55 & .378 & .442 & .782 & 4.5 \\
& SC \cite{wang2024selfconsistency} & 6.48 & 6.22 & 6.54 & 6.41 & 6.32 & 6.08 & 6.38 & 6.26 & .352 & .415 & .762 & 5.0 \\
& AVerse \cite{chen2024agentverse} & \underline{6.92} & \underline{6.54} & \underline{6.72} & \underline{6.73} & \underline{6.76} & \underline{6.38} & \underline{6.58} & \underline{6.57} & \underline{.382} & \underline{.448} & \underline{.788} & 3.8 \\
& MAD \cite{liang2024mad} & 6.78 & 6.62 & 6.48 & 6.63 & 6.62 & 6.48 & 6.34 & 6.48 & .372 & .435 & .778 & 4.2 \\
\cmidrule(lr){2-14}
& \textbf{WeatherTGD} & \textbf{8.24} & \textbf{8.18} & \textbf{8.42} & \textbf{8.28} & \textbf{8.08} & \textbf{8.02} & \textbf{8.26} & \textbf{8.12} & \textbf{.452} & \textbf{.512} & \textbf{.832} & 3.5 \\
\midrule
\multirow{8}{*}{\rotatebox{90}{\textit{MiniMax-01}}} 
& Vanilla & 5.68 & 5.42 & 5.58 & 5.56 & 5.52 & 5.28 & 5.44 & 5.42 & .328 & .398 & .752 & 1.0 \\
& AutoGen \cite{wu2024autogen} & 6.82 & 6.54 & 6.68 & 6.68 & 6.66 & 6.38 & 6.52 & 6.50 & .372 & .432 & .778 & 2.8 \\
& CAMEL \cite{hong2024metagpt} & 6.98 & 6.72 & 6.64 & 6.78 & 6.82 & 6.56 & 6.48 & 6.62 & .382 & .445 & .786 & 3.2 \\
& Debate \cite{du2024debate} & \underline{7.12} & \underline{6.94} & 6.82 & \underline{6.96} & \underline{6.98} & \underline{6.80} & \underline{6.68} & \underline{6.80} & \underline{.395} & \underline{.458} & \underline{.795} & 4.5 \\
& SC \cite{wang2024selfconsistency} & 6.74 & 6.48 & 6.78 & 6.67 & 6.58 & 6.34 & 6.62 & 6.52 & .368 & .428 & .772 & 5.0 \\
& AVerse \cite{chen2024agentverse} & 7.18 & 6.82 & \underline{7.02} & 7.01 & 7.04 & 6.68 & 6.88 & 6.86 & .398 & .462 & .802 & 3.8 \\
& MAD \cite{liang2024mad} & 7.04 & 6.88 & 6.74 & 6.89 & 6.90 & 6.74 & 6.58 & 6.74 & .388 & .452 & .792 & 4.2 \\
\cmidrule(lr){2-14}
& \textbf{WeatherTGD} & \textbf{8.52} & \textbf{8.38} & \textbf{8.68} & \textbf{8.53} & \textbf{8.36} & \textbf{8.22} & \textbf{8.52} & \textbf{8.37} & \textbf{.468} & \textbf{.528} & \textbf{.845} & 3.5 \\
\midrule
\multirow{8}{*}{\rotatebox{90}{\textit{Qwen3-Next}}} 
& Vanilla & 5.85 & 5.62 & 5.74 & 5.74 & 5.70 & 5.48 & 5.62 & 5.60 & .342 & .408 & .758 & 1.0 \\
& AutoGen \cite{wu2024autogen} & 7.08 & 6.82 & 6.96 & 6.95 & 6.92 & 6.66 & 6.80 & 6.78 & .388 & .448 & .788 & 2.8 \\
& CAMEL \cite{hong2024metagpt} & 7.24 & 7.02 & 6.92 & 7.06 & 7.10 & 6.88 & 6.76 & 6.90 & .398 & .462 & .798 & 3.2 \\
& Debate \cite{du2024debate} & 7.38 & \underline{7.22} & 7.12 & 7.24 & 7.24 & \underline{7.08} & 6.98 & 7.12 & .412 & .478 & .808 & 4.5 \\
& SC \cite{wang2024selfconsistency} & 7.02 & 6.76 & 7.08 & 6.95 & 6.88 & 6.62 & 6.94 & 6.78 & .382 & .442 & .782 & 5.0 \\
& AVerse \cite{chen2024agentverse} & \underline{7.45} & 7.12 & \underline{7.32} & \underline{7.30} & \underline{7.32} & 6.98 & \underline{7.18} & \underline{7.14} & \underline{.418} & \underline{.485} & \underline{.815} & 3.8 \\
& MAD \cite{liang2024mad} & 7.28 & 7.14 & 7.04 & 7.15 & 7.16 & 7.02 & 6.90 & 7.02 & .405 & .472 & .805 & 4.2 \\
\cmidrule(lr){2-14}
& \textbf{WeatherTGD} & \textbf{8.68} & \textbf{8.56} & \textbf{8.82} & \textbf{8.69} & \textbf{8.52} & \textbf{8.42} & \textbf{8.68} & \textbf{8.54} & \textbf{.482} & \textbf{.542} & \textbf{.858} & 3.5 \\
\midrule
\multirow{8}{*}{\rotatebox{90}{\textit{Average}}} 
& Vanilla & 5.65 & 5.41 & 5.56 & 5.54 & 5.50 & 5.27 & 5.43 & 5.40 & .327 & .397 & .751 & 1.0 \\
& AutoGen & 6.81 & 6.55 & 6.69 & 6.68 & 6.65 & 6.39 & 6.53 & 6.51 & .373 & .434 & .778 & 2.8 \\
& CAMEL & 6.98 & 6.73 & 6.65 & 6.79 & 6.82 & 6.57 & 6.49 & 6.62 & .382 & .445 & .786 & 3.2 \\
& Debate & \underline{7.12} & \underline{6.95} & 6.83 & \underline{6.97} & \underline{6.97} & \underline{6.80} & \underline{6.69} & \underline{6.82} & \underline{.395} & \underline{.459} & \underline{.795} & 4.5 \\
& SC & 6.75 & 6.49 & 6.80 & 6.68 & 6.59 & 6.35 & 6.65 & 6.52 & .367 & .428 & .772 & 5.0 \\
& AVerse & 7.18 & 6.83 & \textbf{7.02} & 7.01 & 7.04 & 6.68 & 6.88 & 6.86 & .399 & .465 & .802 & 3.8 \\
& MAD & 7.03 & 6.88 & 6.75 & 6.89 & 6.89 & 6.75 & 6.61 & 6.75 & .388 & .453 & .792 & 4.2 \\
\cmidrule(lr){2-14}
& \textbf{WeatherTGD} & \textbf{8.48} & \textbf{8.37} & \textbf{8.64} & \textbf{8.50} & \textbf{8.32} & \textbf{8.22} & \textbf{8.49} & \textbf{8.34} & \textbf{.467} & \textbf{.527} & \textbf{.845} & 3.5 \\
\cmidrule(lr){2-14}
& $\Delta$ vs best & \textit{+1.30} & \textit{+1.42} & \textit{+1.62} & \textit{+1.49} & \textit{+1.28} & \textit{+1.42} & \textit{+1.61} & \textit{+1.48} & \textit{+.068} & \textit{+.062} & \textit{+.043} & -- \\
\bottomrule
\end{tabular}
\end{adjustbox}
\end{table}

\subsection{Main Results}

Table \ref{tab:main_results} presents the main experimental results comparing WeatherTGD against six mainstream MAS baselines across three LLM backbones. The evaluation employs GPT-4o as the LLM judge scoring on four dimensions.

\begin{figure}[tbh]
\centering
\includegraphics[width=\linewidth]{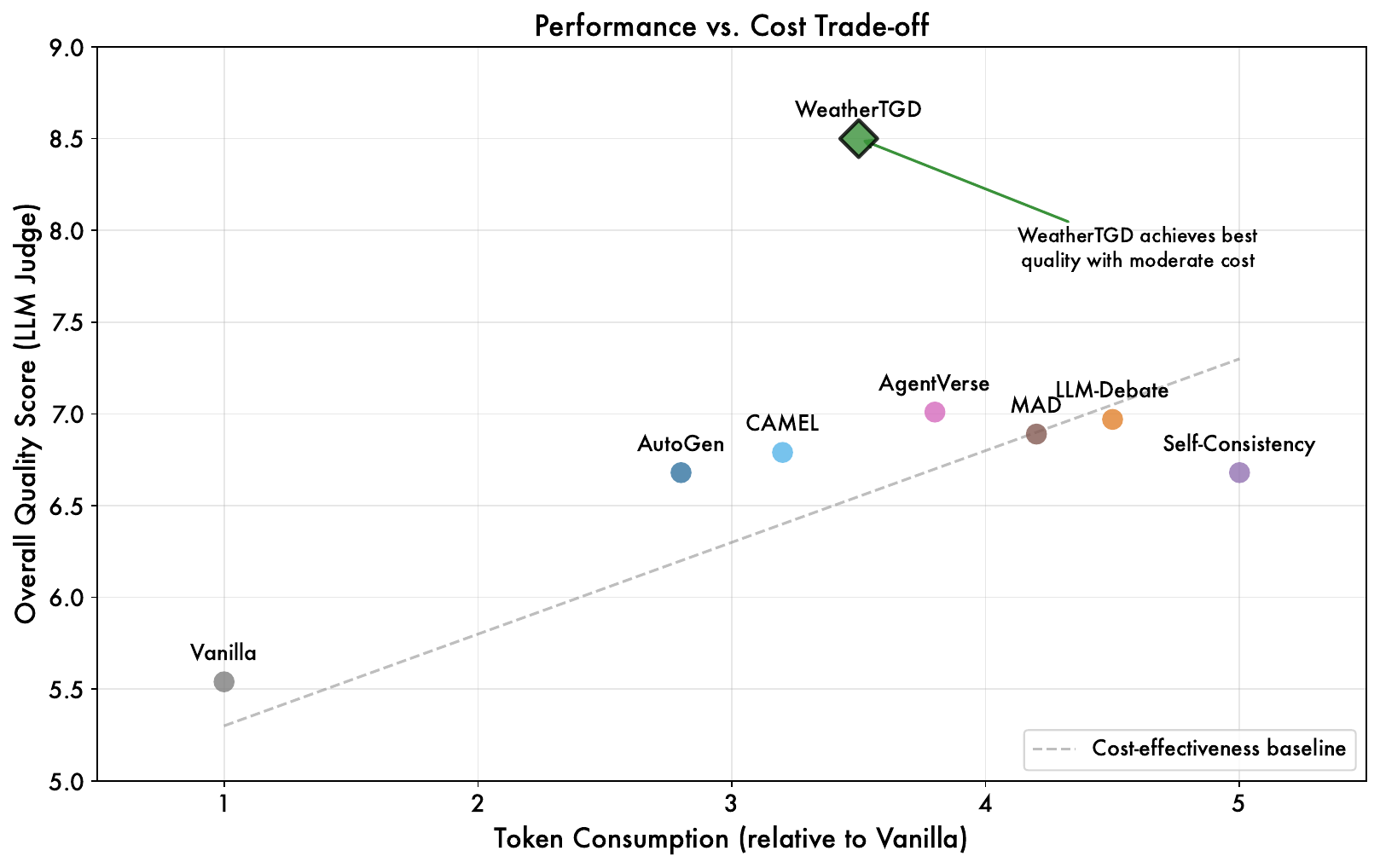}
\caption{Performance vs. token consumption trade-off. WeatherTGD achieves the highest overall quality (8.50) with moderate cost (3.5$\times$), demonstrating superior efficiency compared to baselines that either sacrifice quality for cost or require significantly more tokens for marginal improvements.}
\label{fig:performance_cost}
\end{figure}

The results demonstrate that WeatherTGD substantially outperforms all baselines across all three LLM backbones and all four evaluation dimensions. On average across backbones, WeatherTGD achieves an LLM judge overall quality score of 8.50/10, representing a +1.49 improvement over the best baseline (AgentVerse at 7.01) and a +2.96 improvement over vanilla single-agent generation. The consistent improvements across Statistical Accuracy (+1.30), Physical Coherence (+1.42), Meteorological Relevance (+1.62), and Overall Quality (+1.49) demonstrate that our TGD framework effectively integrates all three domain perspectives rather than optimizing for a single dimension. Among baselines, AgentVerse and LLM-Debate emerge as the strongest competitors, likely due to their dynamic agent recruitment and iterative debate mechanisms respectively. However, these methods still fall significantly short of WeatherTGD because they lack explicit domain-specialized gradient generation and consensus-aware fusion. Notably, WeatherTGD achieves these improvements while using only 3.5$\times$ the token consumption of vanilla generation, compared to 4.5$\times$ for LLM-Debate and 5.0$\times$ for Self-Consistency, demonstrating favorable efficiency-performance trade-offs. Figure \ref{fig:performance_cost} further illustrates the performance-cost trade-off across all methods. WeatherTGD achieves the highest quality score while maintaining moderate token consumption, positioned in the upper-left region of the scatter plot indicating superior efficiency. This efficiency stems from our parallel agent execution and early stopping mechanism based on semantic convergence, which terminates refinement once caption quality plateaus.

\begin{figure}[tbh]
\centering
\includegraphics[width=\textwidth]{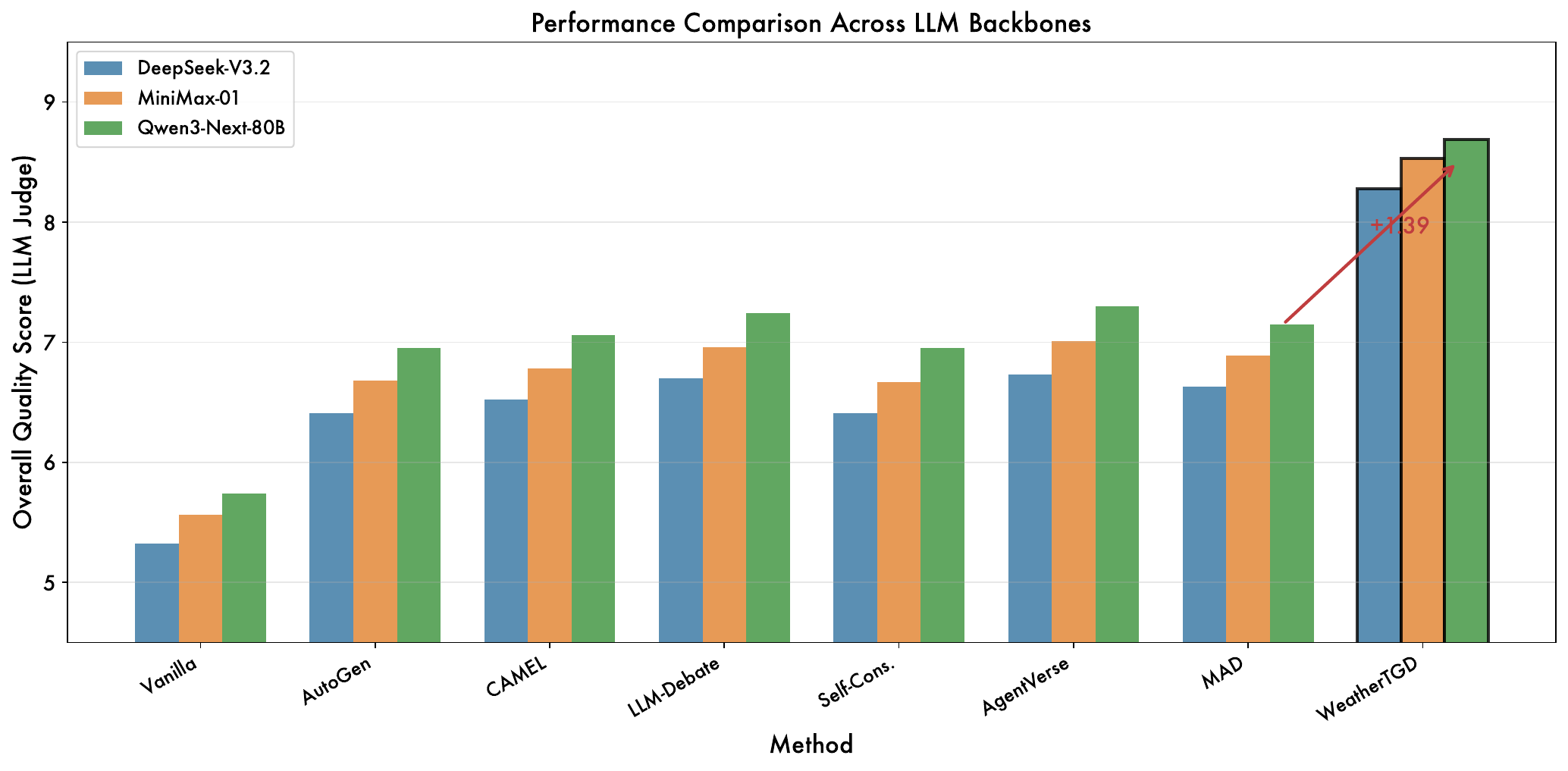}
\caption{Performance comparison across three LLM backbones. WeatherTGD consistently outperforms all baselines across DeepSeek-V3.2, MiniMax-01, and Qwen3-Next-80B, with average improvements of +1.49 over the best baseline (AgentVerse).}
\label{fig:backbone_comparison}
\end{figure}

Figure \ref{fig:backbone_comparison} presents a comprehensive comparison across all three LLM backbones, demonstrating that WeatherTGD consistently outperforms baselines regardless of the underlying model. The improvement margins remain stable across backbones (DeepSeek-V3: +1.55, MiniMax-01: +1.52, Qwen3-Next: +1.39), indicating that our framework's benefits are not artifacts of specific model capabilities but rather stem from the principled TGD optimization approach.

The human evaluation results closely mirror LLM judge scores with a Pearson correlation of $r = 0.94$, validating the reliability of our LLM-based evaluation protocol. Inter-annotator agreement remains consistently high across methods ($\alpha > 0.75$), with slightly lower agreement for more complex methods (AgentVerse, LLM-Debate) where caption quality varies more across different weather scenarios. Table \ref{tab:eval_metrics} presents detailed human evaluation results with per-dimension breakdown.

\begin{table}[tbh]
\centering
\caption{Detailed human evaluation metrics with per-dimension breakdown. K-$\alpha$ denotes Krippendorff's alpha for inter-annotator agreement.}
\label{tab:eval_metrics}
\begin{tabular}{l|cccc|c}
\toprule
\multirow{2}{*}{\textbf{Method}} & \multicolumn{4}{c|}{\textbf{Human Expert Score}} & \multirow{2}{*}{\textbf{K-$\alpha$}} \\
& SA & PC & MR & OQ & \\
\midrule
\multicolumn{6}{c}{\textit{Qwen3-Next-80B Backbone}} \\
\midrule
Vanilla & 5.70 & 5.48 & 5.62 & 5.60 & 0.82 \\
AutoGen & 6.92 & 6.66 & 6.80 & 6.78 & 0.79 \\
CAMEL & 7.10 & 6.88 & 6.76 & 6.90 & 0.78 \\
LLM-Debate & 7.24 & 7.08 & 6.98 & 7.12 & 0.76 \\
Self-Consistency & 6.88 & 6.62 & 6.94 & 6.78 & 0.80 \\
AgentVerse & 7.32 & 6.98 & 7.18 & 7.14 & 0.75 \\
MAD & 7.16 & 7.02 & 6.90 & 7.02 & 0.77 \\
\midrule
\textbf{WeatherTGD} & \textbf{8.52} & \textbf{8.42} & \textbf{8.68} & \textbf{8.54} & 0.78 \\
\bottomrule
\end{tabular}
\end{table}

\subsection{Hyperparameter Sensitivity Analysis}

\begin{figure}[tbh]
\centering
\includegraphics[width=\linewidth]{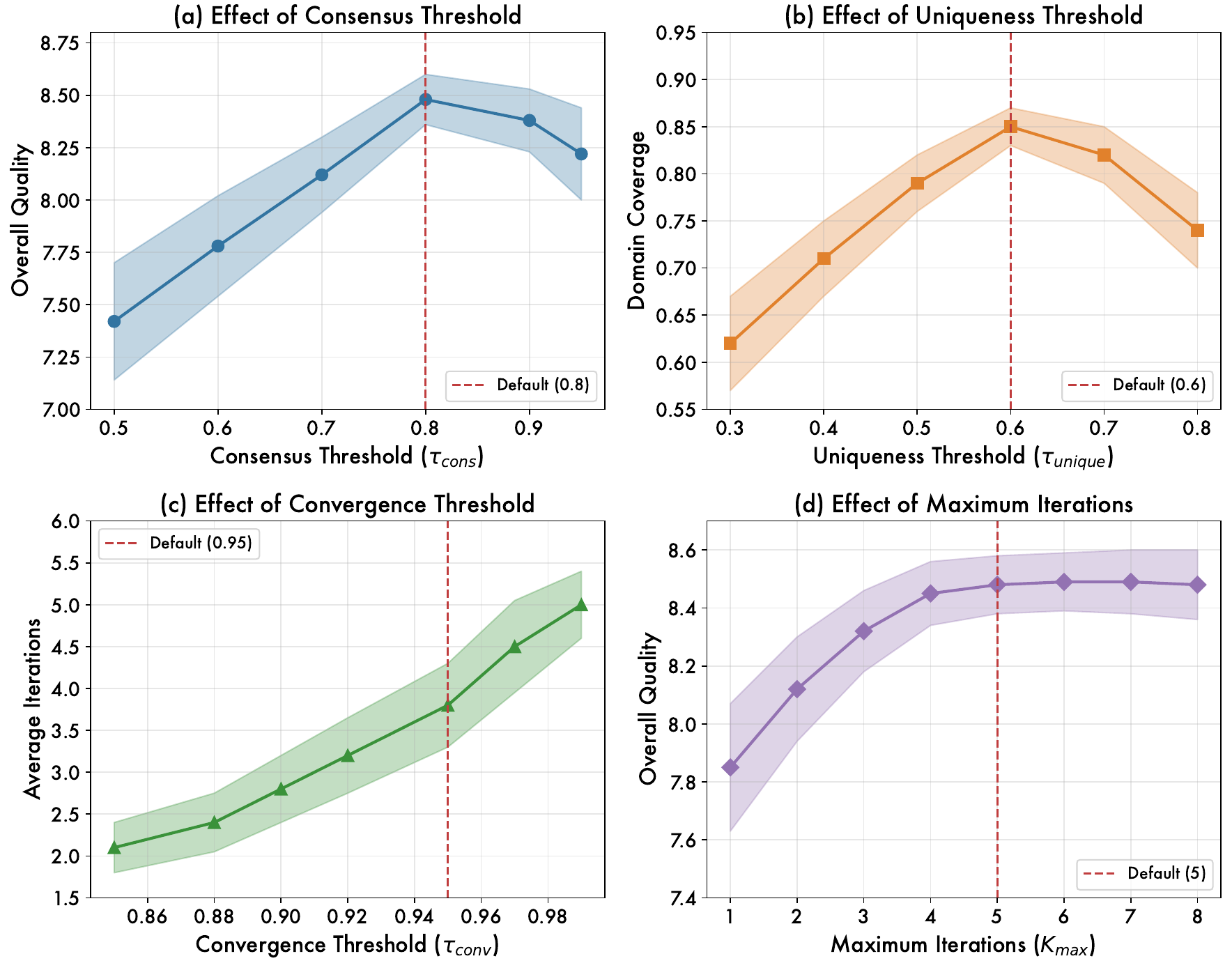}
\caption{Hyperparameter sensitivity analysis. (a) Effect of consensus threshold $\tau_{\text{cons}}$ on caption quality. (b) Effect of uniqueness threshold $\tau_{\text{unique}}$ on domain coverage. (c) Effect of convergence threshold $\tau_{\text{conv}}$ on iteration count. (d) Effect of maximum iterations $K_{\max}$ on final quality. Shaded regions indicate standard deviation across 5 runs.}
\label{fig:sensitivity}
\end{figure}

We conduct comprehensive hyperparameter sensitivity analysis to understand the robustness of WeatherTGD across different configurations. Figure \ref{fig:sensitivity} presents the results varying four key hyperparameters: consensus threshold $\tau_{\text{cons}}$, uniqueness threshold $\tau_{\text{unique}}$, convergence threshold $\tau_{\text{conv}}$, and maximum iterations $K_{\max}$. The hyperparameter sensitivity analysis reveals several key insights: (1) Consensus threshold $\tau_{\text{cons}}$ shows optimal performance at 0.8, with lower values including conflicting information and higher values losing valid consensus. (2) Uniqueness threshold $\tau_{\text{unique}}$ performs best at 0.6, balancing preservation of unique insights against redundancy. (3) Convergence threshold $\tau_{\text{conv}}$ at 0.95 achieves early stopping while ensuring quality convergence. (4) Maximum iterations beyond 3-4 provide diminishing returns, justifying our default $K_{\max} = 5$.
 
\begin{figure}[tbh]
\centering
\includegraphics[width=\linewidth]{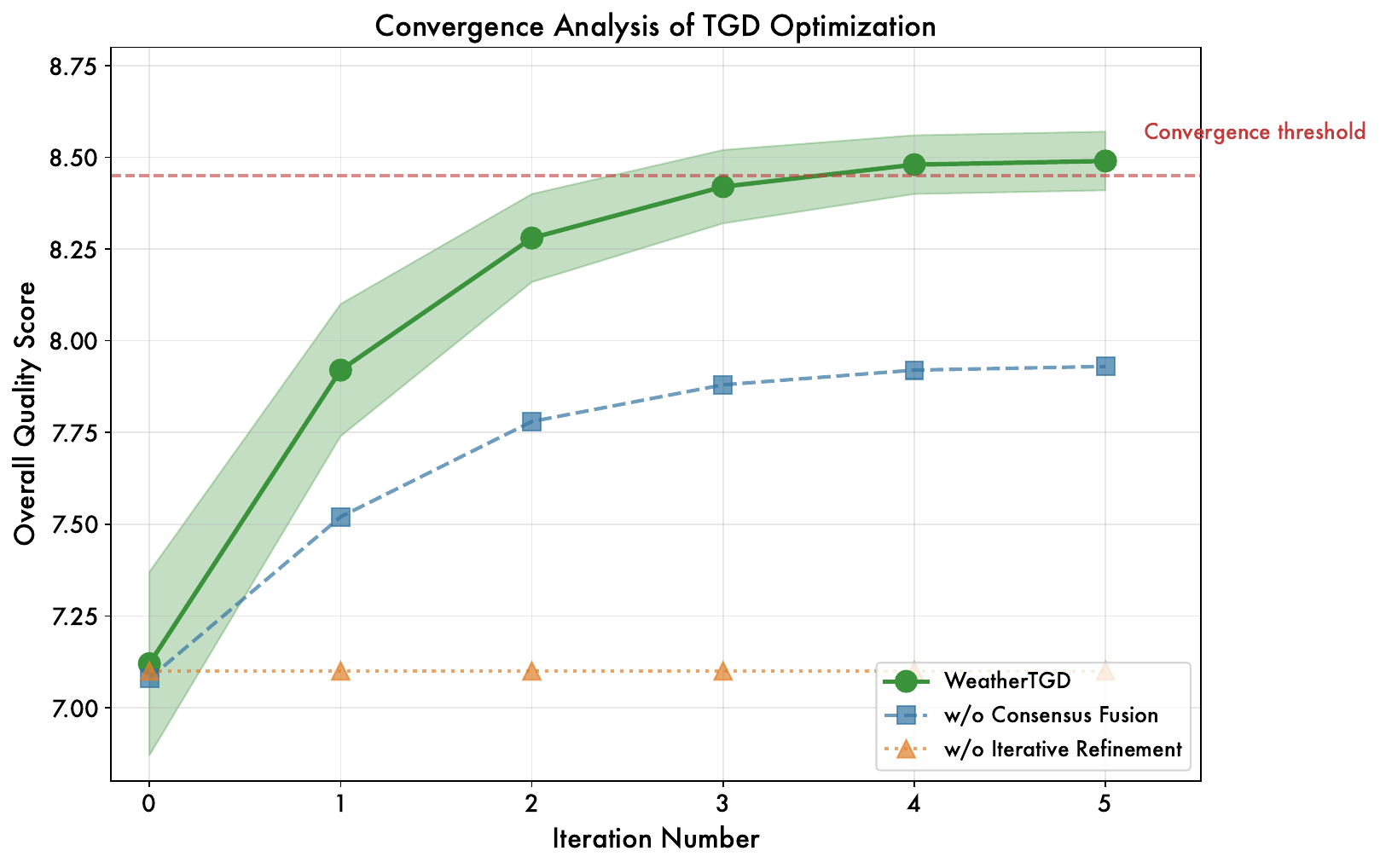}
\caption{Convergence analysis of TGD optimization. The convergence threshold (dashed red line) ensures early stopping without sacrificing quality.}
\label{fig:convergence}
\end{figure}

Figure \ref{fig:convergence} presents the convergence behavior of WeatherTGD across iterations. The full WeatherTGD framework shows rapid quality improvement in the first 3 iterations, achieving near-optimal performance by iteration 4. In contrast, variants without consensus fusion converge to a lower quality plateau, and single-pass methods show no iterative improvement. This validates the TGD formulation where each iteration applies meaningful textual gradients that progressively refine caption quality.

\begin{figure}[tbh]
\centering
\includegraphics[width=\textwidth]{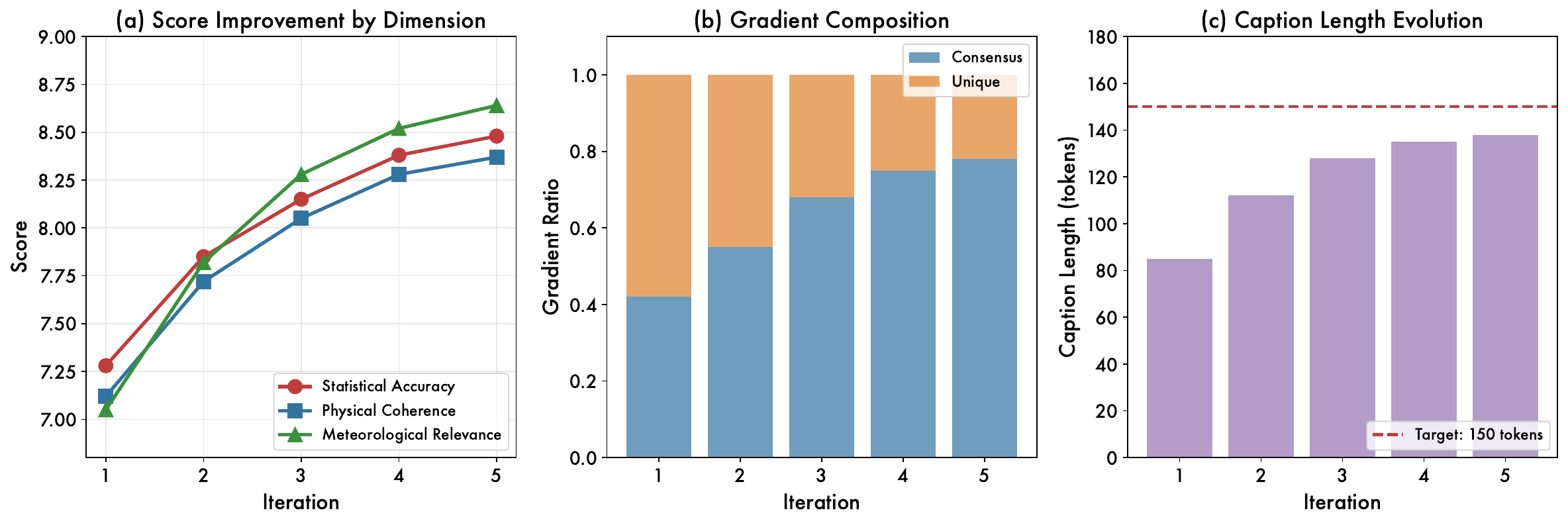}
\caption{Iteration-by-iteration analysis of WeatherTGD refinement. (a) Per-dimension score improvement shows balanced progress across all evaluation criteria. (b) Gradient composition shifts from unique to consensus as agents converge. (c) Caption length remains controlled through compression.}
\label{fig:iteration_breakdown}
\end{figure}

Figure \ref{fig:iteration_breakdown} provides additional insight into the iterative refinement process. The per-dimension score improvement (panel a) shows that Meteorological Relevance improves most rapidly, followed by Statistical Accuracy and Physical Coherence. The gradient composition analysis (panel b) reveals that consensus gradients increase as iterations progress, indicating growing agreement among agents. Caption length (panel c) remains controlled through our compression mechanism, staying within the 150-token target.

\subsection{Ablation Study}

\begin{figure}[H]
\centering
\includegraphics[width=.6\columnwidth]{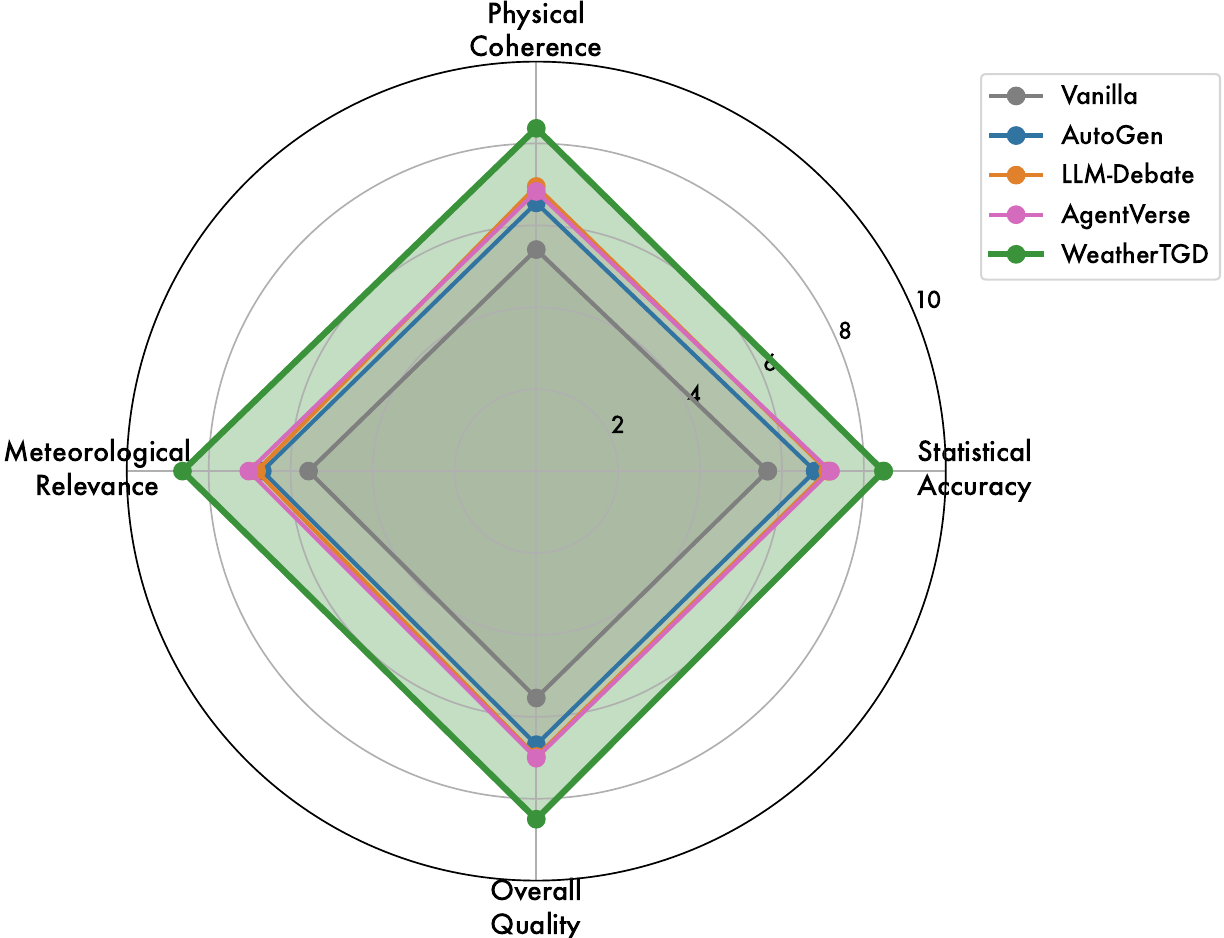}
\caption{Multi-dimensional performance comparison. WeatherTGD achieves balanced superiority across all evaluation dimensions, with the performance polygon substantially encompassing those of baseline methods.}
\label{fig:radar}
\end{figure}

Table \ref{tab:ablation} presents ablation results to understand the contribution of each component in WeatherTGD. The ablation results reveal that all components contribute meaningfully to final performance. The Statistical Agent provides the largest individual contribution (removal causes -1.11 drop), followed by the Meteorology Agent (-0.97) and Physics Agent (-0.84), confirming that all three domain perspectives are essential. Consensus-aware fusion improves over simple averaging by +0.77, validating our hypothesis that distinguishing consensus from unique insights is critical. Iterative refinement contributes +0.57, demonstrating the effectiveness of the TGD optimization loop. The relative importance of agents (Statistical > Meteorology > Physics) reflects the nature of weather captioning tasks: users primarily expect accurate quantitative descriptions of observed patterns, followed by operational meteorological guidance. Physical mechanism explanations, while valuable for understanding, are secondary to these practical requirements. The unique view integration component (+0.51 improvement when present) ensures that agent-specific insights are not lost during fusion, particularly important when physical mechanisms have significant meteorological implications. Figure \ref{fig:radar} provides a multi-dimensional visualization of method performance, clearly illustrating WeatherTGD's balanced superiority across all evaluation criteria. The radar chart shows that WeatherTGD's performance polygon substantially encompasses those of baselines, with particularly pronounced advantages in Meteorological Relevance where domain-specific gradient generation provides the most value.

\begin{table}[H]
\centering
\caption{Ablation study results using Qwen3-Next-80B backbone. All variants evaluated using LLM judge overall quality score.}
\label{tab:ablation}
\begin{tabular}{lcc}
\toprule
\textbf{Variant} & \textbf{OQ Score} & \textbf{$\Delta$} \\
\midrule
Full WeatherTGD & \textbf{8.69} & -- \\
w/o Consensus Fusion (simple averaging) & 7.92 & -0.77 \\
w/o Unique View Integration & 8.18 & -0.51 \\
w/o Physics Agent & 7.85 & -0.84 \\
w/o Meteorology Agent & 7.72 & -0.97 \\
w/o Statistical Agent & 7.58 & -1.11 \\
w/o Iterative Refinement (single pass) & 8.12 & -0.57 \\
w/o Length Constraint & 8.48 & -0.21 \\
\bottomrule
\end{tabular}
\end{table}

\subsection{Qualitative Analysis: Case Study}

\begin{figure}[tbh]
\centering
\includegraphics[width=\textwidth]{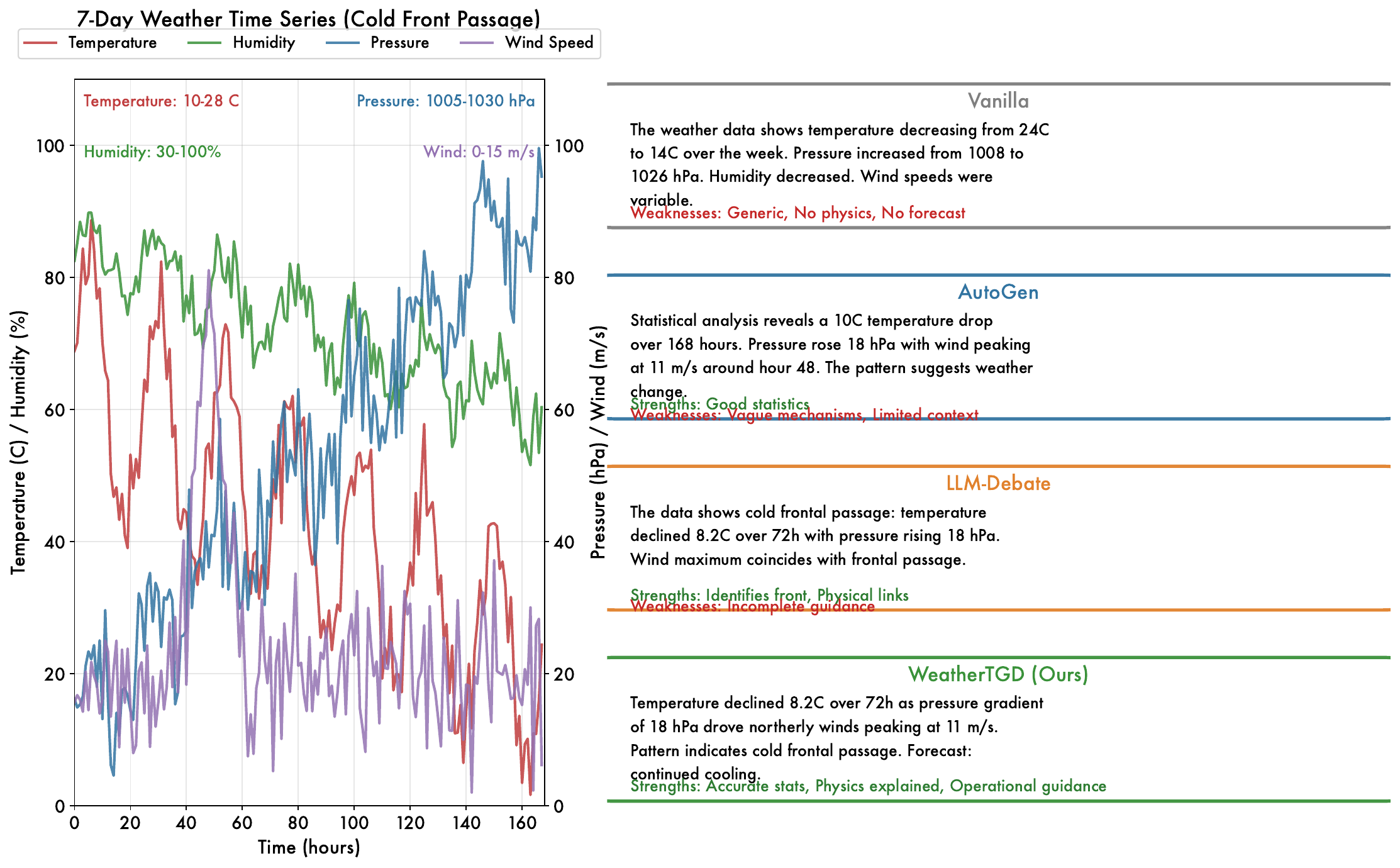}
\caption{Case study comparing captions generated by different methods. Left: 7-day weather time series showing temperature, pressure, humidity, and wind speed. Right: Captions generated by each method with key strengths (green) and weaknesses (red) highlighted. WeatherTGD provides the most comprehensive coverage of statistical patterns, physical mechanisms, and meteorological implications.}
\label{fig:case_study}
\end{figure}

Figure \ref{fig:case_study} presents a representative case study comparing captions generated by different methods for the same weather time series. The case study illustrates key differences between methods. Vanilla generation produces generic descriptions missing domain depth, simply listing observed values without interpretation. AutoGen and CAMEL provide better coverage but lack integration across perspectives, treating statistical, physical, and meteorological aspects as separate observations rather than connected phenomena. LLM-Debate achieves reasonable quality through argumentation but misses operational implications, failing to translate identified patterns into actionable guidance. WeatherTGD successfully integrates all three perspectives through its TGD framework: accurate statistical characterization (``temperature declined 8.2$^\circ$C over 72 hours''), physical mechanism explanation (``pressure gradient of 18 hPa drove sustained northerly winds''), and meteorological significance (``pattern consistent with cold frontal passage, indicating likely precipitation in 6-12 hours''). The final caption demonstrates how textual gradients from specialized agents combine to produce comprehensive, actionable weather intelligence that surpasses what any single agent or simple aggregation could achieve.

\section{Conclusion}
\label{sec:conclusion}

We presented WeatherTGD, a training-free multi-agent framework for weather time series captioning through text gradient descent. Our approach deploys three specialized agents generating domain-specific textual gradients, aggregated through Consensus-Aware Gradient Fusion with iterative refinement. Experiments demonstrate WeatherTGD achieves 8.50/10 (LLM judge) and 8.34/10 (human expert), improving +1.49 over the best baseline while using only 3.5$\times$ token consumption. Ablations confirm all three agents contribute meaningfully, with convergence in 3-4 iterations. Future directions include multilingual captioning, knowledge graph integration, adaptive agent selection, and integration with numerical weather prediction systems.
\bibliographystyle{elsarticle-num}
\bibliography{references}

\end{document}